\documentclass[conference]{IEEEtran}
\IEEEoverridecommandlockouts
\usepackage[all]{nowidow}
\usepackage{nameref}

\usepackage{subcaption}

\usepackage[americanresistors,americaninductors]{circuitikz}
\usepackage{tikz}
\usepackage{siunitx}
\usetikzlibrary{decorations.pathreplacing,calligraphy, decorations.pathmorphing,patterns}
\usetikzlibrary{shapes.geometric, arrows}
\usepackage{amsmath}
\usepackage{graphicx}
\usepackage[colorlinks=true, allcolors=blue]{hyperref}

\usepackage{placeins}

\usepackage{cite}
\usepackage{amsmath,amssymb,amsfonts}
\usepackage{algorithmic}
\usepackage{graphicx}
\usepackage{textcomp}
\usepackage{xcolor}
\def\BibTeX{{\rm B\kern-.05em{\sc i\kern-.025em b}\kern-.08em
    T\kern-.1667em\lower.7ex\hbox{E}\kern-.125emX}}

\usepackage{cleveref}

\Crefname{equation}{Eq.}{Eqs.}
\Crefname{figure}{Fig.}{Figs.}
\Crefname{tabular}{Tab.}{Tabs.}

\DeclareSIUnit\inHg{inHg}

\relax

\bibstyle{IEEEtran}
\bibdata{references}
\gdef \@abspage@last{3}

\begin{document}

\title{Macro-Scale Electrostatic Origami Motor}

\makeatletter
\newcommand{\linebreakand}{%
  \end{@IEEEauthorhalign}
  \hfill\mbox{}\par
  \mbox{}\hfill\begin{@IEEEauthorhalign}
}
\makeatother

\author{
  Alex S. Miller\textsuperscript{1}*, Leo McElroy\textsuperscript{2}, Jeffrey H. Lang\textsuperscript{3}\\
  \\
\textsuperscript{1}Dept. of Aeronautics and Astronautics, \textit{Massachusetts Institute of Technology}, Cambridge, USA\\
\textsuperscript{2}\textit{Independent Researcher}, Cambridge, USA\\
\textsuperscript{3}Dept. of Electrical Eng. and Computer Sci.
    \textit{Massachusetts Institute of Technology}, Cambridge, USA\\
    *Corresponding author
}

\maketitle

\begin{abstract}
Origami structures have been an active area of research due to their high volume-to-mass ratio, packability, and shape adaptability. Previously-developed actuated foldable structures have either embedded linear actuators into the structure, or have attached non-folding rotary motors to the exterior structure. Further, in prior work, actuators directly embedded in the structure of folding media all contributed to linear or folding motion, not to continuous rotary motion. On the macro-scale there has not yet been a folding continuous rotary actuator. This paper details the development and testing of the first macro-scale origami rotary motor that can be folded flat, and then unfurled to operate. Using corona discharge for torque production, the prototype motor achieved an expansion ratio of over 3.8:1, reached a top speed of 1720\,rpm when driven at -30\,kV, and exhibited a maximum motor torque over 0.25\,mN\,m with an active component torque density of 0.11\,N\,m\,kg$^{-1}$.
\end{abstract}

\iffalse
\fi

\section*{Introduction}

Motors are a critical technology underpinning the conversion of energy in modern society. The wide variety of applications for motors in energy conversion, actuation and locomotion drive an ecosystem of widely-varying motor sizes, electromagnetic topologies and power levels. 

Origami has enabled structures to change form \cite{hawkes_programmable_2010} \cite{miyashita_untethered_2015}, to pack into smaller sizes, and to achieve a high volume-to-mass ratio \cite{meloni_engineering_2021}. Actuation and locomotion in foldable structures remains difficult due to the necessity of thin materials. Existing origami devices have been actuated by external fields \cite{miyashita_untethered_2015}\cite{ze_spinning-enabled_2022}, by non-folding attached rotary motors \cite{sung_foldable_2015}\cite{carlson_rebound_2020}\cite{felton_method_2014}, by cable actuation\cite{ochalek_adaptive_2024}, or by linear actuators on the origami structure such as shape memory alloy \cite{felton_method_2014}\cite{hu_review_2021}, resistive heaters \cite{park_field-programmable_2025}, or super-coiled conductive polymer actuators \cite{yan_origami-based_2023}. Folding structures have contributed to useful advances in space structures, rovers, and satellites; since space systems are launched in a fixed volume, origami enables small stowed packages to transform into large structures \cite{meloni_engineering_2021}\cite{turner_review_2016}\cite{gong_design_2022}. A collapsible motor allows for a torque per volume ratio that increases as the motor is unfurled into the deployed position. Additionally, the low volume of two-dimensional building materials can allow foldable systems to be transported in low-volume containers and assembled in-situ\cite{zhu_large-scale_2024}. Foldable motor construction techniques also allow for potential integration into existing foldable structural architectures\cite{ze_spinning-enabled_2022}\cite{zhang_plug_2023}\cite{lee_self-locking_2024}. 

Electric fields rather than magnetic fields are the clear choice for implementing a foldable rotary actuator, since electrodes are far easier to fabricate in a 2D form than magnets, coils, or cores. Electric-field machines favor high surface area of material, whereas magnetic field machines favor high volume of material. Electrostatic motors and other electric-field machines have captured the attention of actuator designers since the 1740s \cite{jefimenko_electrostatic_2011}. Electric-field machines may be built using variable capacitance, electrostatic induction, permanent electrets, or corona discharge as the active mechanism of torque production \cite{jefimenko_electrostatic_2011}.  An electrostatic motor made using DNA origami techniques was recently demonstrated on the nano scale, but this motor was not readily collapsible after construction \cite{pumm_dna_2022}. In this paper, we design and demonstrate a corona motor that is the first macro-scale origami motor, and the first collapsible continuous rotary motor.

\section*{Results}

\begin{figure*}[t!]
        \centering
        
        \begin{subfigure}[b]{0.3\textwidth}
            \phantomsubcaption  % invisible caption
            \label{fig:full_assembly}
        \end{subfigure}%
        \begin{subfigure}[b]{0.3\textwidth}
            \phantomsubcaption
            \label{fig:stowed_side}
        \end{subfigure}%
        \begin{subfigure}[b]{0.3\textwidth}
            \phantomsubcaption
            \label{fig:stowed_top}
        \end{subfigure}%
        \begin{subfigure}[b]{0.3\textwidth}
            \phantomsubcaption
            \label{fig:unstowed_side}
        \end{subfigure}%
        \begin{subfigure}[b]{0.3\textwidth}
            \phantomsubcaption
            \label{fig:unstowed_bottom}
        \end{subfigure}%

        \includegraphics[width=1\linewidth]{media/Figure_1.png}
        \caption{Macro-scale electrostatic origami motor. (a) Assembled electrostatic origami motor with components labeled. The body of the electrostatic origami motor unfurles from 16 \si{\mm} to 62 \si{\mm}, not including external shaft height, which stays constant through the expansion and may be an arbitrary length. The stowed configuration is shown in side view and top view in (b) and (c) respectively, and the stable unfurled configuration is shown in side view and bottom view in (d) and (e) respectively. The stowed configuration is maintained by external force; after unfurling, the motor is stable in the deployed position.}
        \label{fig:pictures_of_motor_and_folding}
\end{figure*}

\subsection*{Motor Design} \label{sec:design}

The motor in this work, shown in \Cref{fig:pictures_of_motor_and_folding}, is designed as a corona discharge motor, which uses discharge from pointed sections on the stator to deposit charge on the rotor, creating shear force between the stator and the rotor \cite{jefimenko_electrostatic_2011}\cite{krein_analysis_1995}. This mechanism is depicted in \Cref{fig:corona-motor-mechanicanism_summary}, and contextualized in a cylindrical motor geometry with notable dimensions in \Cref{fig:corona-motor-mechanicanism}. This motor was designed as a corona discharge motor to allow for asynchronous direct-current (DC) operation, simplifying test electronics.

To achieve a folding origami motor with radial electric fields requires two nested collapsible cylindrical patterns. There are many possibilities for achieving collapsible cylindrical origami, including a Kresling \cite{kresling_natural_2008}, Yoshimura \cite{yoshimura_mechanism_1955}\cite{von_karman_buckling_1941}, Tachi-Miura \cite{tachi_one-dof_2010}, or Waterbomb cylinder tesselation \cite{lang_twists_2017}. For this application, a Kresling pattern is selected due to its well-known geometric design constraints and the varying stability possibilities that the pattern offers \cite{lang_twists_2017}\cite{zang_kresling_2024}. Particularly, the Kresling pattern has high resistance to axial bending which aids bearing design, enabling a more precise electrostatic gap than other origami patterns. The Kresling pattern in this work was designed to be monostable in the deployed position, but future designs could use bistable Kresling patterns to enable more sophisticated deployment schemes. A photograph of the assembled origami motor in this work is shown in \Cref{fig:full_assembly}.  The collapse and stowing action is shown in \Crefrange{fig:stowed_side}{fig:unstowed_bottom}, with the system achieving a stowed height of \SI{16}{\mm} and a deployed height of \SI{62}{\mm}.

One of the cylinders, the rotor, is rigidly fixed onto central shafts at both ends. The other cylinder, the stator, nests outside of the rotor with bearings on the shaft at both ends to support radial and axial forces. To enable precise fabrication of electrodes onto their flexible surfaces, the rotor and stator are fabricated with flexible printed-circuit boards (flexPCBs); paper and conductive tape were effective in the prototyping process. To enable operation with a direct-current power supply, the motor presented in this paper is designed as a corona discharge motor. Thus, the stator has electrodes that are excited by a high-voltage power supply, and the rotor is completely passive.

The geometry of the rotor and stator are defined by the radius of the inscribed circle of the Kresling cylinder, $R$, the number of segments around the circumference, $N$, and $\theta_0$, the angle between the base and diagonal of the parallelogram unit cell. A diagram of a single unit cell is shown in \Cref{fig:unit-cell}. The side length of each unit is calculated as
\begin{equation}
    a = 2R \sin\left(\frac{\pi}{N}\right) .
\end{equation}
The angle between the consecutive polygons in the folded configuration \cite{lang_twists_2017} \cite{zang_kresling_2024} is
\begin{equation}
    \theta_{\mathrm{max}} = \pi - \frac{2\pi }{N} - \theta_0\ .
\end{equation}
This sets the height of the unit cell \cite{lang_twists_2017} \cite{zang_kresling_2024} at
\begin{equation}
    h = R \sqrt{2\left[\cos(\theta_0) - \cos(\theta_{max})\right]}\ . 
\end{equation} 

The stator and rotor support electrode patterns which enable the device to function as a motor. The stator employs strip electrodes on the inside of the flexPCB, positioned on the long diagonals which stick in towards the rotor. These electrodes are positioned on the inner-most surfaces of the stator to minimize the gap from the stator to the rotor, and to act as a sharp surface which concentrates the electric field, thus promoting corona discharge.

The rotor has a series of pads on top of an insulating substrate which collect corona discharge from the stator. These pads are placed on the short vertical sides which stick out most prominently from the rotor exterior. These pads are segmented to prevent shorting from different areas of the motor, since both the rotor and stator are slanted. The rotor and stator electrode patterns are replicated $N$ times circumferentially and $M$ times axially. Tabs are added to the top and bottom of each pattern which attach to rigid components and terminal electrodes, and perforations are used to encourage clean folds. The rotor and stator sizes are chosen so that the rotor can spin freely within the stator in the fully deployed configuration.

\begin{figure*}[t!]
        \begin{subfigure}[b]{0.3\textwidth}
            \phantomsubcaption
            \label{fig:corona-motor-mechanicanism_summary}
        \end{subfigure}%
        \begin{subfigure}[b]{0.3\textwidth}
            \phantomsubcaption
            \label{fig:corona-motor-mechanicanism}
        \end{subfigure}%
        \begin{subfigure}[b]{0.3\textwidth}
            \phantomsubcaption  % invisible caption
            \label{fig:forward-and-reverse-mode}
        \end{subfigure}%
        \centering
        \includegraphics[width=1\linewidth]{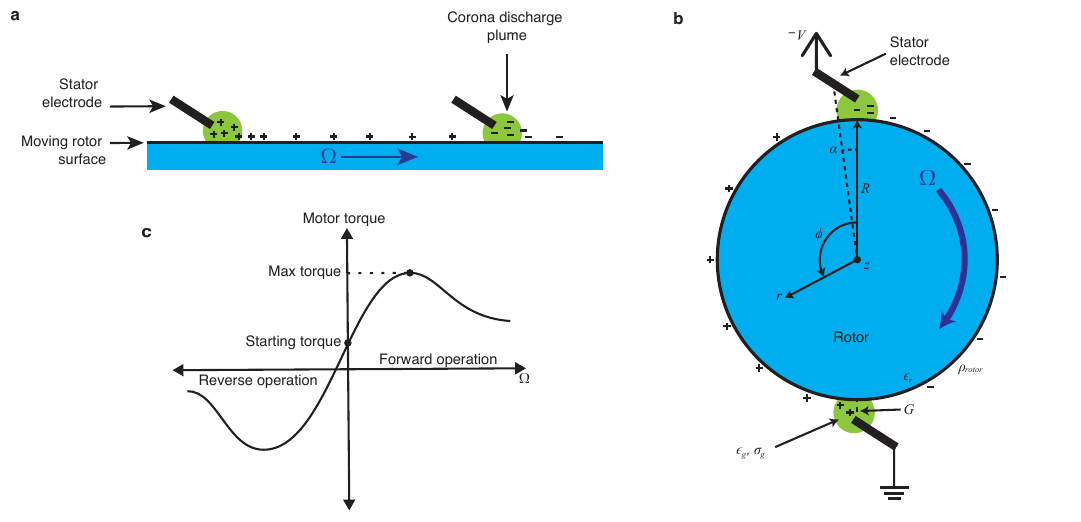}
        \caption{The physical mechanism behind the macro-scale origami motor. (a) A graphical representation of the corona discharge mechanism which causes motor torque with physical parameters labeled. The rotor is pictured in blue, the two stator electrodes are depicted with thick dark lines outside of the rotor surface, and ionized air is shown in green. (b) This mechanism occurs on a cylindrical stator model, with parameters labeled. (c) An representation of corona motor torque as a function of speed, showing the forward and reverse modes of operation. The motor is biased in starting direction due to the $\alpha$ angular electrode offset.}
        \label{fig:placeholder}
\end{figure*}

The design is implemented in JSON-PCB, an open-source tool for designing printed circuit boards with scripts \cite{leomcelroy_leomcelroyjson-pcb_2025} \cite{mcelroy2022svg}, seen in \Cref{fig:JSON-PCB}. The circuit generating script is written in JavaScript and targets the JSON-PCB format, which can then be viewed in the web-based editor and exported in Gerber format for manufacturing. The main structures have dimensions shown in \Cref{tab:dims}. The $R$ dimension parameters of the stator and rotor are chosen to facilitate rotary function, maintaining a consistent gap between the rotor and stator. The $\theta_0$ parameter is chosen to be the same across the rotor and stator to facilitate similar folding behavior during the nesting fold, and to be monostable in the deployed configuration. The maximum value of $N$ for maintaining consistent folding dynamics, for a given $\theta_0$ and $R$, is set by strain limitations in the substrate, since the Kresling pattern is not rigid-foldable. The stator was designed with an even $N$, since corona motors are excited using even numbers of phases, with $N=8$ leading to sharp corona discharge features. The rotor was designed with an $N$ as high as possible to make the deployed shape close to that of a cylinder, minimizing gap variance. The height of the motor was chosen as $M=2$, to demonstrate the nesting fold design across multiple units. To allow the PCBs to join in a cylindrical shape, with a taped overlap region, the circuit boards are manufactured with a horizontal extent of $N+1$. The rotor has copper on the outer surface (front) and the stator has corona electrodes on the inner surface (back), and terminals on both the outer and inner surfaces. Both the stator and the rotor are manufactured on bare polyimide with no coverlay. Ommitting coverlay on both circuit boards allows for a more compact stowed device. Since the stator electrodes abut a fold line, avoiding stress concentrations at this fold line is critical to the longevity of the device after multiple fold cycles. Omitting coverlay on the stator circuit board avoids creating stress concentrations along these fold lines, since coverlay would force the substrate between the coverlay and copper region to bend at a tighter radius. The manufactured flexPCB rotor and stator are depicted in \Cref{fig:rotor} and \Cref{fig:stator} respectively.

\begin{table}[t!]
\centering
\caption{Key dimensions in stator and rotor.}
\begin{tabular}{|c|c|c|}
\hline
           & Stator Dimension              & Rotor Dimension               \\ \hline
$R$        & 41\,mm                         & 30\,mm                         \\ \hline
$\theta_0$ & \ang{57} & \ang{57} \\ \hline
$N$        & 8                             & 10                            \\ \hline
$M$        & 2                             & 2                             \\ \hline
\end{tabular}
\label{tab:dims}
\end{table}

Rigid 3D-printed parts connect the rotor and stator to their respective bearing and shaft surfaces. At both ends of the rotor there is an a short shaft to transmit motor output power. Both sides of the stator connected to rigid liner pieces, which then fit into an end cap. These end caps are fitted with sapphire ring jewel bearings which enable low-friction motor operation. The ring jewels were picked to have olive-shaped holes to allow for acceptable performance with high misalignment. Assembly details are described in the \nameref{sec:assembly} section. A cross-section rendering showing the interfaces between the rigid and origami parts is depicted in \Cref{fig:cross-section}.

\begin{figure*}[t!]

        \begin{subfigure}[b]{0.3\textwidth}
            \phantomsubcaption  % invisible caption
            \label{fig:unit-cell}
        \end{subfigure}%
        \begin{subfigure}[b]{0.3\textwidth}
            \phantomsubcaption
            \label{fig:JSON-PCB}
        \end{subfigure}%
        \begin{subfigure}[b]{0.3\textwidth}
            \phantomsubcaption
            \label{fig:rotor}
        \end{subfigure}%
        \begin{subfigure}[b]{0.3\textwidth}
            \phantomsubcaption
            \label{fig:stator}
        \end{subfigure}%
        \begin{subfigure}[b]{0.3\textwidth}
            \phantomsubcaption
            \label{fig:rotor_wrapping}
        \end{subfigure}%
        \begin{subfigure}[b]{0.3\textwidth}
            \phantomsubcaption
            \label{fig:rotor_completed}
        \end{subfigure}%
        \begin{subfigure}[b]{0.3\textwidth}
            \phantomsubcaption
            \label{fig:motor_stackup_assembly}
        \end{subfigure}%
        \begin{subfigure}[b]{0.3\textwidth}
            \phantomsubcaption
            \label{fig:motor_assembled_assembly_figure}
        \end{subfigure}%
        \begin{subfigure}[b]{0.3\textwidth}
            \phantomsubcaption
            \label{fig:cross-section}
        \end{subfigure}%
        \begin{subfigure}[b]{0.3\textwidth}
            \phantomsubcaption
            \label{fig:high_speed_footage}
        \end{subfigure}%

        \centering
        \includegraphics[width=1\linewidth]{media/Figure_3.png}
        \caption{The design and assembly process of a macro-scale origami motor. (a) Key dimensions in the mechanism Kresling unit cell fold design. (b) The flexible circuit boards for the rotor and stator are designed in JSON-PCB, an open source script-based PCB design tool especially suited to parametric designs \cite{leomcelroy_leomcelroyjson-pcb_2025}\cite{mcelroy2022svg}. The (c) rotor and (d) stator printed circuit boards. To assemble the rotor structure, in (e) the rotor is wrapped around a rigid arbor and affixed with double sided tape, with a full-length alignment shaft. Then the connections are glued, the edges are trimmed, and partial-length shafts are installed, leading to the finished rotor in (f). (g) Assembly process of the stator end caps, stator and rotor. (h) Finished motor assembly. (i) Cross section of origami motor computer rendering with components labeled. The Kresling structure was generated using a CAD script \cite{hanson_controlling_2024} and adjusted to system constraints. (j) Motor speed was measured using a high-speed camera, with a tracking fiducial.}
        \label{fig:placeholder}
\end{figure*}

\subsection*{Motor Performance}
The motor was connected to a high-voltage DC power supply, with a pair of opposite motor electrodes connected to power and ground and with a current limit of \SI{0.5}{\milli\ampere} for safety. All data were collected with the central axis of the motor perpendicular to gravity. Movies of motor rotation were taken using a high-speed camera (Edgertronic SC1 Monochrome with Nikon 24-85\,\si{\milli\meter} f/2.8-4.0 AF-D) lens recording at 500\,fps. These frame data were processed using the Tracker Video Analysis and Modeling Tool \cite{brown_tracker_2026} to track a small dot on the rotor. The rotor fiducial dot in a frame taken from the high-speed camera is shown in \Cref{fig:high_speed_footage}. Angular velocity and acceleration were calculated from position measurements of the dot in each frame.  The motor was able to self start at voltages exceeding \SI{-27}{\kilo\volt} in amplitude. For these voltages, time-series data are shown in \Cref{fig:timeseries}, where the motor was powered on at a fixed DC voltage, then powered off.  All data in \Cref{fig:data_main_paper} were taken on August 17\textsuperscript{th}, 2026 in Cambridge, MA, in a lab environment with a barometric pressure of \SI{29.86}{\inHg}, humidity of \SI{63}{\percent}, and temperature of \SI{22.2}{\celsius} (measured with Taylor 1735 2-node weather station). The \SI{-30}{\kilo\volt} trial only was filmed during steady state and spindown. All trials show oscillations in speed due to the rotor periodically moving axially towards the stator end cap. When started from rest, the motor almost always started in the same direction. When the motor was manually started in the other direction, it operated in a reverse-direction mode with a lower-speed, as seen in \Cref{fig:reverse_mode_data}, further supporting the model developed in \nameref{sec:analysis} and shown in \Cref{fig:forward-and-reverse-mode}. 

\begin{figure*}[t!]
        \begin{subfigure}[b]{0.3\textwidth}
            \phantomsubcaption  % invisible caption
            \label{fig:timeseries}
        \end{subfigure}%
        \begin{subfigure}[b]{0.3\textwidth}
            \phantomsubcaption
            \label{fig:spindown}
        \end{subfigure}%
        \begin{subfigure}[b]{0.3\textwidth}
            \phantomsubcaption
            \label{fig:speed-torque}
        \end{subfigure}%
        \begin{subfigure}[b]{0.3\textwidth}
            \phantomsubcaption
            \label{fig:steady-state-speed}
        \end{subfigure}%
        \begin{subfigure}[b]{0.3\textwidth}
            \phantomsubcaption
            \label{fig:torque-voltage}
        \end{subfigure}%
        \begin{subfigure}[b]{0.3\textwidth}
            \phantomsubcaption
            \label{fig:mechanical-power}
        \end{subfigure}%
        \begin{subfigure}[b]{0.3\textwidth}
            \phantomsubcaption
            \label{fig:current-voltage}
        \end{subfigure}%
        \begin{subfigure}[b]{0.3\textwidth}
            \phantomsubcaption
            \label{fig:efficiency}
        \end{subfigure}%
        \begin{subfigure}[b]{0.3\textwidth}
            \phantomsubcaption
            \label{fig:force-displacement}
        \end{subfigure}%

        \centering
        \includegraphics[width=1\linewidth]{media/Figure_4.png}
        \caption{Performance characterization of origami motor. (a) Speed data time-series collected with high-speed camera. (b) Torque from bearing friction and air resistance during three unpowered spin-down tests with a corresponding fitted load curve. (c) Torque-speed curve of active motor components tested at three different voltages with the load subtracted. (d) Observed steady state speed as a function of drive voltage plotted with a linear fit. (e) Steady state torque by motor active components as a function of voltage compared to model. (f) Steady state mechanical output power as a function of input voltage. (g) Measured current through a shunt resistor as a function of voltage. (h) Calculated efficiency of motor as a function of voltage. (i) Mechanical restoring force of rotor, stator and full motor system measured for various motor lengths.}
        \label{fig:data_main_paper}
\end{figure*}

From the time-series data in \Cref{fig:timeseries}, the output torque can be calculated as 
\begin{equation}
    \mathcal{T}_\mathrm{output}   = I_{\mathrm{rotor}}\frac{\mathrm{d}^2\theta}{\mathrm{d} t^2},
\end{equation}
where $I_{\mathrm{rotor}}$ is the moment of inertia and $\theta$ is the rotation angle of the spinning rotor. In the prototype tested here, $I_{\mathrm{rotor}} =$ \SI{4.24e-06}{\kg\meter\squared}. The total output torque, $\mathcal{T}_\mathrm{output}$, is the sum of the active motor torque, $\mathcal{T}_\mathrm{motor}$, and the load torque of bearing friction and aerodynamic drag, $\mathcal{T}_\mathrm{load}$ where
\begin{equation} \label{eq:motor_torque_equation}
    \mathcal{T}_\mathrm{output} = \mathcal{T}_\mathrm{motor} - \mathcal{T}_\mathrm{load}.
\end{equation}

After power is disconnected from the motor, $\mathcal{T}_\mathrm{output} = -\mathcal{T}_\mathrm{load}$. Therefore, the spin-down phase was used to characterize the bearing and windage loads in the motor. In \Cref{fig:spindown}, the spin-down load torque is plotted and a quadratic load curve is fitted. The trial 1 data in \Cref{fig:spindown} was similar to trials 2 and 3 but had a small instability observed during the spindown and thus was excluded from the quadratic fit. These data show the bearing and windage loads to be time invariant across multiple trials, including those from other testing days such as in \Cref{fig:data_supplement_paper}. This technique characterizes dynamic friction, but not static friction, since the system is in motion during spin-down.  Static friction depends on the lubrication state of the bearings, the stationary time of the system, and the condition of the rotor shaft; across testing it has been observed between $0.02$\,--\,$0.05$\,\si{\milli\newton\meter}.

From the characterization of $\mathcal{T}_\mathrm{load}$, $\mathcal{T}_\mathrm{motor}$ was calculated for various driving voltage levels, shown in \Cref{fig:speed-torque}. The motor is able to run at speeds where the torque is above the load curve during speed-up, and reaches a steady state when the motor speed intersects the load curve. Oscillations in speed due to mechanical factors, such as occasional axial rotor movement, were present. Minor oscillations around the steady state appear near the load curve intersection. These measurements were plotted alongside the model developed in \nameref{sec:analysis}, Eqs. (\ref{eq:torque-speed-eqn}) and (\ref{constitutive_equation}) with the parameters $\sigma = $ \SI{4}{\nano\siemens\per\m}, $R=$ \SI{30}{\mm}, $L=$ \SI{54}{\mm},  $G=$ \SI{3.5}{\mm}, $\epsilon_g = \epsilon_0$, $\epsilon_r = \epsilon_0$, $V_{\mathrm{onset}}=$ \SI{-9.0}{\kilo\volt}, and $\alpha = $ \SI{0.10}{\radian}. This value of $\sigma$ is similar to that found by Krein \cite{krein_analysis_1995}, and the value for $V_{\mathrm{onset}}$ is consistent with the size of the gap, $G$. This model shares key characteristics with these data, having comparable zero-speed torque and maximum torque.

In addition to the measurements made during the full speed-up and slow-down, measurements were also recorded once the motor reached steady state. These data were taken by first starting the motor at \SI{-30}{\kilo\volt}, waiting for the motor to reach steady state, taking a video measurement, then decreasing the magnitude of excitation by \SI{0.5}{\kilo\volt}, waiting for the motor to reach steady state, and taking a new measurement. This process was repeated until the motor was no longer able to run continuously, which occurred below \SI{-19}{\kilo\volt} amplitude. This experimental process allowed access to lower voltage operating points than would be accessible starting from rest due to the static friction of the bearing being greater than the dynamic friction. During this testing, the motor was run continuously for a period of over \SI{120}{\minute}. These measurements are shown in \Cref{fig:steady-state-speed}.

Using the load curve from the spin-down tests, these data were transformed to represent the steady state motor torque, $\mathcal{T}_\mathrm{motor}$ at the same voltages as shown in \Cref{fig:torque-voltage}. These measurements are compared to the model developed in \nameref{sec:analysis}, Eqs. (\ref{eq:torque-speed-eqn-max}) and (\ref{constitutive_equation}), with the same parameters as used in \Cref{fig:speed-torque} and are similar in shape and magnitude.  The mechanical power $P_\mathrm{motor,mech} = \Omega \mathcal{T}_{\mathrm{motor}}$ is calculated and shown in \Cref{fig:mechanical-power}.

Current measurements were recorded using a multimeter (Fluke 117) through a \SI{9.850}{\kilo\ohm} shunt resistor, with the motor stationary, on the same day as the rest of the motor experimental data. Two trials of current data are shown in \Cref{fig:current-voltage}, matching the shape of the constitutive relation in \Cref{constitutive_equation}. The efficiency, $\eta = P_\mathrm{motor,mech} / P_\mathrm{in}$, where $P_\mathrm{in} = VI$ is calculated using the averaged current between both measurement trials, with fitted speed and torque data. The calculated efficiency is shown in \Cref{fig:efficiency}. The motor is less than \SI{2}{\percent} efficient for all input voltage ranges; this is likely due to the relatively narrow electrode blades, which have additional sharp surfaces which are not pointed towards the rotor. Techniques for increasing motor efficiency are discussed in the \nameref{sec:discussion} section. Additional motor test data are plotted in \Cref{fig:data_supplement_paper}; these data show similar results, with minor differences in $\sigma$ and $V_{\mathrm{onset}}$, likely due to environmental conditions.

Mechanically, the rotor was folded more than 50 times without visible damage. The stator, and correspondingly, the whole motor, was folded more than 20 times without any sign of damage. Earlier stator versions, manufactured with coverlay and a polyimide overlay, were only able to withstand 5 cycles of folding before significant cracking along the stator electrodes; this problem was addressed by manufacturing the stator version in this paper without coverlay, as described in \nameref{sec:design}. The force needed to compress the rotor, stator and combined motor structure was measured as a function of displacement with a digital force gauge (Extech 475040), and shown in \Cref{fig:force-displacement}. Full compression of the whole structure was attained with \SI{16}{\newton} of force. The rotor is \SI{54}{\milli\meter} in the unfurled state, and is able to compress to a height of \SI{6}{\milli\meter}; which is smaller than required to nest within the folded stator; this is shown in the dotted section of \Cref{fig:force-displacement}. In \Cref{fig:force-displacement}, the rotor displacement was offset by \SI{8}{\milli\meter} to align with the stator compression, for comparison. When linearized over the first \SI{2}{\milli\meter} displacement from the unfurled state, the spring constant for the rotor is \SI{335}{\newton\per\meter}; with a \SI{8.9}{\gram} rotor mass, this suggests a resonant frequency ($f = \frac{1}{2 \pi}\sqrt{k/m}\,$) around \SI{30.9}{\hertz}, or 1857\,rpm. Calculated with the same method, over the first \SI{2}{\milli\meter} displacement from the unfurled state, the spring constant for the stator is \SI{70}{\newton\per\meter}; with a \SI{31.0}{\gram} stator mass, this suggests a resonant frequency around \SI{7.6}{\hertz}, or 450\,rpm. Given that both of these resonant frequencies are in or near the range of motor operation, it is possible
that they may be excited during motor testing, but these frequencies are both faster than the oscillations seen in \Cref{fig:timeseries}. The rotor also exhibits an imbalance due to the overlapping portion where the cylindrical form is glued together; this effect is observed at low speeds in the spindown test in \Cref{fig:spindown} and might add energy into resonant modes for some motor operating points. 

The assembled origami motor was able to deploy from a stowed body height of \SI{16}{\mm} to a deployed body height of \SI{62}{\mm}, resulting in a body expansion ratio of over 3.8:1. The motor reached a maximum speed of 1720\,rpm when driven at \SI{-30}{\kilo\volt} as shown in \Cref{fig:steady-state-speed} and produced \SI{0.25}{\milli\newton\meter} of motor torque, $\mathcal{T}_\mathrm{motor}$, as shown in \Cref{fig:torque-voltage}. The maximum output torque, $\mathcal{T}_\mathrm{output}$, as defined in \Cref{eq:motor_torque_equation}, was measured at \SI{0.10}{\milli\newton\meter}, as seen in the difference between the motor torque and load curves in \Cref{fig:speed-torque}. The motor torque value, $\mathcal{T}_\mathrm{motor}$, which is independent of the load, is used for torque density calculations. With a deployed volume of \SI{0.00033}{\cubic\meter}, and stowed volume of \SI{0.000084}{\cubic\meter}, the prototype motor exhibited a volumetric torque density of \SI{0.77}{\newton\meter\per\cubic\meter} in the deployed configuration, which increased to \SI{3.0}{\newton\meter\per\cubic\meter} in the stowed configuration. Considering a total system mass of \SI{39.9}{\gram}, the motor demonstrated a torque density of \SI{0.0055}{\newton\meter\per\kg} and power density of \SI{1.1}{\watt\per\kilo\gram}, and considering only the active circuit board components with a mass of \SI{0.75}{\gram} for the rotor flexPCB and \SI{1.40}{\gram} for the stator flexPCB, the motor demonstrated a torque density of \SI{0.11}{\newton\meter\per\kg} and power density of \SI{23.25}{\watt\per\kilo\gram}.

\section*{Discussion} \label{sec:discussion}
This paper presents the first macro-scale origami rotary motor. The active components of this motor are able to be fabricated from flat sheets, then folded into a 3D structure. Additionally, the assembled motor can be collapsed and deployed repeatedly. This work shows that electrostatic rotary motors can be made from flexible materials. Implementation from materials that may be flat-packed may enable bulk material transport and in-situ assembly\cite{zhu_large-scale_2024}. Devices like this may be used in soft robotics applications where a robot must fit into a small packed volume and then operate in a deployed state.  Additionally, devices like this may enable collapsible versions of ultralight flying vehicles \cite{shen_sunlight-powered_2024}, integration of an additional mode of actuation into existing folding origami structures for medicine \cite{ze_spinning-enabled_2022}, manipulation \cite{zhang_plug_2023}, or programmable structures \cite{lee_self-locking_2024}, or adding actuation to other non-origami compact deployable structures such as solar sails\cite{wilkie_advanced_nodate}.

While this work is useful in demonstrating the merits of electrostatic folding machines, there are several notable areas for future work. Future folding motors may be made with thinner materials, lower-profile rigid components and different folding patterns that enable a lower-volume folded state towards the goal of increasing volumetric torque density. The current prototype, made of polyimide with rigid end caps achieves a stowed body thickness of \SI{16}{\milli\meter} and a stowed volumetric torque density of \SI{3.0}{\newton\meter\per\cubic\meter}. By folding the design out of \SI{10}{\micro\meter} thin origami paper with \SI{1}{\micro\meter} thick electrodes deposited through a mask, and substituting the current plastic rigid components for methyl-cellulose-stiffened paper with adhered bearings and shaft mounts, we expect that a design very similar to this prototype architecture made out of these different materials could reach thicknesses under \SI{250}{\micro\meter}, leading to a torque to density of \SI{192}{\newton\meter\per\cubic\meter}. Thus, the expansion ratio acts as a torque density multiplier. In bringing electrostatic machines to a technology level competitive with magnetic machines, multiplicative gains from reducing gap size, using dielectric fluid, and implementing new control schemes may be implemented independently \cite{ludois_macroscale_2022}; an expansion ratio for electrostatic machines may act as a further torque density multiplier independent from these other factors.

New designs may also be made with tighter gaps between the rotor and stator by having more folded unit cells, and thus making them closer to cylindrical. Additionally, taller machines can be made by stacking more than two unit cells, yielding linearly more torque, as shown in \nameref{sec:analysis} with the linear dependence of torque in \Cref{eq:torque-speed-eqn} on $L$, and thus $M$. Future designs may incorporate sharper stator electrode features to concentrate local electric fields and to initiate corona discharge at lower voltages, with a larger $\alpha$ to increase starting torque, as per the model in \Cref{eq:torque-speed-eqn}. Additionally, future versions may implement more stator pole pairs. The motor in this paper was subject to mechanical oscillation, such as those seen in the \SI{-29}{\kilo\volt} trial in \Cref{fig:timeseries_supp}, due to axial freedom of motion as well as structural compliance; new designs may incorporate domed stator bearings facing the rotor to lightly pre-load the motor axially without adding significant friction, with the goal of dampening these oscillations. Adding such axial bearings would also improve performance when the motor axis is parallel to the axis of gravity. Future efforts may optimize the dimensions and parameters in \Cref{tab:dims} to allow for bistability or more efficient operation, and may also incorporate latching mechanisms to allow the motor to be deployed as part of a robotic system. 

The motor presented here operates using corona discharge with the primary goal of producing torque with a direct-current supply, and demonstrating the concept of an origami rotary actuator. However, the motor in this work runs around \SI{-30}{\kilo\volt}, posing a safety limitation. Additionally, corona discharge is inefficient, difficult to scale, and depends on the properties of the gap medium, including the operating environment conditions. To open up possibilities for variable scale, higher torque and more potential applications, future folding motors may be designed around a variable-capacitance \cite{jefimenko_electrostatic_2011}\cite{bollee_electrostatic_1969}, electrostatic induction \cite{melcher_continuum_1981}\cite{krein_analysis_1995}\cite{nagle_electric_2005}, or separately-excited \cite{ludois_macroscale_2022} topology, driven with multiple phases of high-voltage alternating current. At the macro-scale, with newly-developed amplifier designs for electrostatics applications \cite{jurgiel_high-voltage_2026}\cite{mitchell_pocket-sized_2022}, it is possible to create the multi-phase alternating-current excitation to drive a variable capacitance or induction machine. At the micro and nano scales, these topologies lead magnetics in performance \cite{nagle_electric_2005}\cite{livermore_high-power_2004}. Combining the advantages of nano-scale electrostatic working gaps with collapsibility opens up possibilities of further multiplicative torque density gains and new manufacturing possibilities from 2D materials. The form and specifications of the other actuators with lightweight or soft materials has been listed for comparison in \Cref{tab:comparison-table-motors}. While the motor in this work has lower torque density specifications than several state of the art devices, the motor in this work is the only currently-known macro-scale foldable continuous rotary actuator. As previously noted, new manufacturing techniques, smaller scales, use of a dielectric fluid, and more efficient electrical topology are expected yield several orders of magnitude \cite{ludois_macroscale_2022} of improvement to the actuator in this work.

Future generations of foldable motors offer applications worthy of further exploration, especially for high-speed lower-torque applications. For instance, linear swimming motion by spinning a Kresling-based origami structure has been previously achieved through external fields \cite{ze_spinning-enabled_2022}; if implemented using a variable capacitance or electrostatic induction architecture, a motor similar to the one in this work could be used to achieve swimming motion through a dielectric fluid without use of external fields. Future versions of ultralight flying vehicles \cite{shen_sunlight-powered_2024} may be made foldable with structures similar to the one in this work. Future origami rotary actuators may be added to space structures that require deployability such as solar sails, or small satellites. Foldable robots requiring continuous rotary motion may benefit from the integration of origami actuators into the structure rather than affixing external actuators or using external fields.  

This work describes the first architecture for a deployable, crushable rotary motor made out 2D materials. Development of a flexible rotary actuator enables future actuator systems to achieve leading volumetric torque density through stowage and enables new categories of integrated flexible actuator systems.

\section*{Methods}
\label{sec:methods}
\subsection*{Analysis}
\label{sec:analysis}

A corona motor is an electrostatic machine where electric discharge from sharp features on the stator causes charge transport in ionized gas towards the rotor; this charge transport causes electrostatic repulsion, resulting shear force and rotary motion. For making simple models, corona motors may be considered a subset of electrostatic induction machines, where the gap conductivity is higher than the rotor conductivity and the excitation is constant. Rotating induction machines in a constant electric field were first characterized by von Quincke \cite{quincke_ueber_1896}; later, Melcher presented a general solution and discussion of the dynamics of electrostatic induction machines  \cite{melcher_continuum_1981}. To use the induction-motor analogy for a corona motor, previous studies have estimated an effective conductivity in the gap to model the ionized air \cite{krein_analysis_1995}. Compared to traditional electrostatic induction motors, corona motors such as the one constructed here usually have discharge blades set at an angle, where the corona discharge originates from the edge of the blade, but the field is centered around the middle of the blade structure. The method presented here accounts for this asymmetry. The effect of the blade asymmetry angle on starting torque has been measured experimentally \cite{hattori_fundamental_1992}, but the method in this work describes it analytically. A graphical depiction of the model with the physical parameters and the electromechanical parameters is shown in \Cref{fig:corona-motor-mechanicanism}. This model is appropriate when only two opposing terminals are excited; if more terminals are used, then a different form of the potential would be needed. 

In cylindrical coordinates, where $r$ is the distance from the center, $z$ is the displacement along the motor shaft, and $\phi$ is the azimuthal angle, the electric potential outside the rotor can be modeled as 
\begin{equation}
    \Phi_{\mathrm{a}} = \operatorname{Re}\left(-E_0re^{j\phi} + \frac{\hat{A}}{r}e^{j\phi}\right), 
\end{equation}
where $E_0$ is the background electric field and $\hat{A}$ is a complex coefficient representing dipole charging of the rotor to be derived from the boundary conditions. Inside the rotor, the potential can be modeled as 
\begin{equation}
    \Phi_{\mathrm{b}} = \operatorname{Re}\left(-\hat{B}re^{j\phi}\right), 
\end{equation}
where $\hat{B}$ is a complex coefficient to be derived from the boundary conditions. On the surface of the rotor, where $r=R$, continuity of the electric potential dictates $\Phi_{\mathrm{a}}|_{r=R} = \Phi_{\mathrm{b}}|_{r=R}$, yielding a relationship between $\hat{A}$ and $\hat{B}$ where
\begin{equation}
    E_0 = \frac{\hat{A}}{R^2} + \hat{B}\ . \label{pcons}
\end{equation}
Using the electric fields $E_a = -\nabla\Phi_a$ and $E_b = -\nabla\Phi_b$,  charge conservation on the rotor surface yields
\begin{equation}
    \hat{r}\cdot \left( \sigma E_a\right)|_{r=R} + \left(\frac{\partial}{\partial t} + \Omega\frac{\partial}{\partial \phi}\right)\rho_{\mathrm{rotor}}  
    + J_{\mathrm{c}}e^{j(\phi -\alpha)} =  0 \, , \label{qcons}
\end{equation}
where $\Omega$ is the rotor angular velocity, $\sigma$ is the conductivity around the gap, $\rho_{\mathrm{rotor}}$ is the rotor surface charge, and $J_{\mathrm{c}}e^{j(\phi-\alpha)}$ is the surface current density, periodic around the rotor, caused by the corona discharge. This boundary condition uses both $J_{\mathrm{c}}$ and $\sigma E_0$ as modes of charge transport, modeling the charge-relaxation behavior of the rotor using $\sigma$ and the charging asymmetry using $J_{\mathrm{c}}$. The boundary condition may also be set up such that the charge relaxation term is $\hat{\phi}\cdot \left( \sigma E_a\right)|_{r=R}$; this gives the same torque result.  The charge on surface of the rotor at $r=R$ may be expressed as
\begin{equation}\label{rho_equation}
\rho_{\mathrm{rotor}} = \operatorname{Re}\left\{\left[\epsilon_g\left(E_0+\frac{\hat{A}}{R^2}\right) - \epsilon_r\hat{B}\right]e^{j\phi}\right\}, 
\end{equation}
where $\epsilon_g$ is the permittivity in the gap and $\epsilon_r$ is the permittivity of the rotor. 

In steady state where $\frac{\partial\rho_{\mathrm{rotor}}}{\partial t}=0$, (\ref{qcons}) can be written as
\begin{multline}
   \sigma\left(E_0 + \frac{\hat{A}}{R^2}\right) + 
   j\Omega\left[\epsilon_g\left( E_0 + \frac{\hat{A}}{R^2} \right) - \epsilon_r\hat{B}\right] \\
   + J_{\mathrm{c}}e^{-j\alpha}  =  0 \, , \label{qcons2}
\end{multline}
Combining (\ref{pcons}) and (\ref{qcons2}) yields
\begin{equation}\label{A_solved}
    \hat{A} = R^2 \frac{-E_0\left[\sigma+j\Omega(\epsilon_g-\epsilon_r)   \right] - J_{\mathrm{c}}e^{-j\alpha}}{\sigma+j\Omega(\epsilon_g+\epsilon_r)}.
\end{equation}
Substituting (\ref{pcons}) and (\ref{A_solved}) into (\ref{rho_equation}), the charge on the surface of the rotor at $r=R$ may be calculated as
\begin{multline}
    \rho_{\mathrm{rotor}} =
     \frac{-  (\epsilon_{g}+\epsilon_{r})}{{1+\Omega^{2} \tau^{2}} } \frac{J_{\mathrm{c}}}{\sigma} \left[ \cos(\phi-\alpha )  + \Omega \tau \sin(\phi-\alpha)\right] \\ 
     -\frac{  2 \epsilon_r}{{1+\Omega^{2} \tau^{2}} } E_0 \left[ \cos(\phi)  + \Omega \tau \sin(\phi)\right]  ,
\end{multline}
where $\tau$ is a time constant, $\tau = \frac{\epsilon_r + \epsilon_g}{\sigma_g}$. The value of the electromagnetic stress tensor may be evaluated and integrated on the surface of the rotor to calculate rotor torque as 
\begin{multline} \label{eq:torque-speed-eqn}  
    \mathcal{T}_{\mathrm{motor,offset}} = R\int_{0}^{2\pi}  \epsilon_g E_{a,r}|_{r=R}E_{a,\phi}|_{r=R} LRd\theta
    \\
    = 2\pi LR^2    \; \frac{ 2\frac{\epsilon_r\epsilon_g}{\epsilon_g+\epsilon_r}E_0^2\Omega\tau + \epsilon_gE_0\frac{J_{\mathrm{c}}}{\sigma}\left[\sin{(\alpha)} + \Omega\tau\cos{(\alpha)}\right]}{1 + \Omega^2\tau^2},
\end{multline}
where $E_{a,r} = \hat{r} \cdot E_{a}$, $E_{a,\phi} = \hat{\phi} \cdot E_{a}$, $R$ is the rotor radius and $L$ is the rotor axial length. The angular offset $\alpha$ gives rise to a starting torque. In most cases, the motor will start in the same direction from rest, due to this offset, and proceed to a stable equilibrium speed where the load torque equals the motor torque. If the electric field is applied to a rotor already spinning in reverse, it is possible to kickstart the motor into the reverse mode, which yields a lower peak torque, but still operates. These states are shown in \Cref{fig:forward-and-reverse-mode}. The maximum torque is also a function of this asymmetry and is given by
\begin{multline}\label{eq:torque-speed-eqn-max}
    \mathcal{T}_{\substack{\mathrm{motor}\\ \mathrm{max}}} = \pi LR^2 \left[\frac{2 \epsilon_r \epsilon_g E_0^2  }{\epsilon_r + \epsilon_g} + \epsilon_g E_0\frac{J_{\mathrm{c}}}{\sigma}  \;  \left(1 +\sin(\alpha) \right)\right].
\end{multline}
In this case, the ``background" electric field is dependent on the size of the stator. A constitutive equation for an ``effective" electric field suitable for estimating torque can be written as 
\begin{equation} \label{constitutive_equation}
    E_{0,\mathrm{eff}} \approx\frac{J_{\mathrm{c}}}{\sigma} \approx \frac{V - V_\mathrm{onset}}{2R+2G},
\end{equation}
where $V$ is the driving voltage and $V_\mathrm{onset}$ is a correction term for the voltage at which discharge begins. The $V_\mathrm{onset}$ offset depends on atmospheric conditions and the gap. For example, under standard conditions, a breakdown voltage of $\approx$ \SI{3}{\kilo\volt\per\mm} would be typical.

\subsection*{Assembly}
\label{sec:assembly}
All 3D printed parts are printed on a Stratasys J5 MediJet printer with dissolvable supports. For the rotor assembly, the flexPCB rotor is adhered to rigid components at both ends using double-sided tape. This taping is performed by first wrapping the tape around the rigid rotor attachments, then rolling the rotor around them, with a temporary stabilizing shaft as shown in \Cref{fig:rotor_wrapping}. The stabilizing shaft is then removed and the two end-shafts (\SI{3}{\milli\meter} diameter, \SI{13}{\milli\meter} and \SI{20}{\milli\meter} long) are pre-fit into an interference fit on the rotor assembly, as shown in \Cref{fig:rotor_completed}. There is one unit of overlap in the rotor, which is filled with double sided tape. Cyanacrylate adhesive is added near the taping surface to further increase rigidity. 

For the stator assembly, the flexPCB stator is adhered to the rigid liner component at both ends with double-sided tape using the same rolling technique as the rotor. On the stator, the electrode overhang is trimmed completely one one side, and the non-copper overhang are trimmed on the other side; this leaves terminal electrodes that protrude the stator liner, as shown on the assembled stator in \Cref{fig:motor_stackup_assembly}. A ring jewel bearing is then fitted into each stator end cap. 

Next, the full motor is assembled as shown in \Cref{fig:motor_stackup_assembly}. Ring jewel bearings are fitted into each stator end cap to handle radial loads. From bottom to top, the following components are stacked: a stator end cap, the stator with a liner ring, one thrust bearing, the rotor assembly with the plastic bushing down, one more thrust bearing, then the remaining stator end cap. All bearings are the SwissJewel R317.0 model \cite{noauthor_ring_nodate} olive-shaped ring jewel. Depending on the folding precision achieved, additional thrust bearings may be added to further constrain the rotor if needed. Finally, the bearings are lubricated with synthetic watch lubricant (Moebius 9020), dispensed using a 31-gauge syringe. The completed assembly is shown in \Cref{{fig:motor_assembled_assembly_figure}}, with a cross-section detailed view in \Cref{fig:cross-section}.

\section*{Author Contributions}
A.M. and L.M. designed the motor, developed assembly methods and performed experiments. A.M. and J.L. performed analysis. A.M analyzed data. A.M. wrote the manuscript. All authors read and approved the final manuscript.

\section*{Acknowledgments}
%Omitted for review.
A.M. and L.M. acknowledge Erik Demaine and the MIT Origami Folding Algorithms class for learning, inspiration, and guidance. The team thanks Jim Bales for lending high-speed videography and lighting equipment. The authors thank the T.J. Rodgers RLE Laboratory for their helpful equipment and advice.

\section*{Funding}
This material is based upon work supported by the NSF Graduate Research Fellowship under Grant No. 2141064 and by the Fannie and John Hertz Foundation. The funders played no role in study design, data collection, analysis and interpretation of data, or the writing of this manuscript. 

\section*{Competing Interests}
J.L. declares no financial or non-financial competing interests. A patent application (63/904,244), titled ``Collapsible Electrostatic Devices and Related Systems, Structures, and Methods'', now pending, has been filed by MIT on October 23\textsuperscript{rd} 2025, with A.M. and L.M. listed as inventors, assigned to MIT. The application covers the concept and methods of an origami motor developed in this work.

%For patents and patent applications, disclosure of the following information is requested: patent applicant (whether author or institution), name of inventor(s), application number, status of application, specific aspect of manuscript covered in patent application.

\section*{Code Availability}
The underlying code and design files for this study are available in \url{https://github.com/alexmi11/origami-motor} and can be accessed via this link \url{https://doi.org/10.5281/zenodo.22851506}.

\section*{Data Availability}
The datasets generated and analyzed during the current study are available at \url{https://github.com/alexmi11/origami-motor}, which is archived at \url{https://doi.org/10.5281/zenodo.22851506}.

\vspace{-0.095cm}
\FloatBarrier
%\newpage
\hypersetup{
    hidelinks % Makes links invisible but still clickable
}

\end{document}